\documentclass{article}

    \PassOptionsToPackage{numbers}{natbib}

\usepackage[final]{neurips_2022}

\usepackage[utf8]{inputenc} 
\usepackage[T1]{fontenc}    
\usepackage{hyperref}       
\usepackage{url}            
\usepackage{booktabs}       
\usepackage{amsfonts}       
\usepackage{nicefrac}       
\usepackage{microtype}      
\usepackage{xcolor}         
\usepackage[pdftex]{graphicx}
\usepackage{mathtools}
\usepackage{amsmath}
\usepackage{wrapfig,lipsum,booktabs}
\usepackage{verbatim}

\title{Unsupervised Object Representation Learning using Translation and Rotation Group Equivariant VAE}

\author{%
  Alireza Nasiri \\
  Simons Machine Learning Center\\
  New York Structural Biology Center\\
  \texttt{anasiri@nysbc.org} \\
   \And
   Tristan Bepler \\
  Simons Machine Learning Center\\
  New York Structural Biology Center\\
  \texttt{tbepler@nysbc.org}
}

\begin{document}

\maketitle

\begin{abstract}
In many imaging modalities, objects of interest can occur in a variety of locations and poses (i.e. are subject to translations and rotations in 2d or 3d), but the location and pose of an object does not change its semantics (i.e. the object's essence). That is, the specific location and rotation of an airplane in satellite imagery, or the 3d rotation of a chair in a natural image, or the rotation of a particle in a cryo-electron micrograph, do not change the intrinsic nature of those objects. Here, we consider the problem of learning semantic representations of objects that are invariant to pose and location in a fully unsupervised manner. We address shortcomings in previous approaches to this problem by introducing TARGET-VAE, a translation and rotation group-equivariant variational autoencoder framework. TARGET-VAE combines three core innovations: 1) a rotation and translation group-equivariant encoder architecture, 2) a structurally disentangled distribution over latent rotation, translation, and a rotation-translation-invariant semantic object representation, which are jointly inferred by the approximate inference network, and 3) a spatially equivariant generator network. In comprehensive experiments, we show that TARGET-VAE learns disentangled representations without supervision that significantly improve upon, and avoid the pathologies of, previous methods. When trained on images highly corrupted by rotation and translation, the semantic representations learned by TARGET-VAE are similar to those learned on consistently posed objects, dramatically improving clustering in the semantic latent space. Furthermore, TARGET-VAE is able to perform remarkably accurate unsupervised pose and location inference. We expect methods like TARGET-VAE will underpin future approaches for unsupervised object generation, pose prediction, and object detection. Our code is available at \url{https://github.com/SMLC-NYSBC/TARGET-VAE}.
\end{abstract}


\section{Introduction}

In many imaging modalities, objects of interest are arbitrarily located and oriented within the image frame. Examples include airplanes in satellite images, galaxies in astronomy images \cite{lintott2008galaxy}, and particles in single-particle cryo-electron microscopy (cryo-EM) micrographs \cite{sigworth2016principles}. However, neither the location nor the rotation of an object within an image frame changes the nature (i.e. the semantics) of the object itself. An airplane is an airplane regardless of where it is in the image, and different rotations of a particle in a cryo-EM micrograph are still projections of the same protein. Hence, there is great interest in learning semantic representations of objects that are invariant to their locations and poses. 

In general, unsupervised representation learning methods do not recover representations that disentangle the semantics of an object from its location or its pose. Popular unsupervised deep learning methods for images, such as variational autoencoders (VAE) \cite{kingma2013auto}, usually use encoder-decoder frameworks in which an encoder network is learned to map an input image to an unstructured latent variable or distribution over latent variables which is then decoded back to the input image by the decoder network. However, the unstructured nature of the latent variables means that they do not separate into any specific and interpretable sources of variation. To achieve disentanglement, methods have been proposed that encourage independence between latent variables \cite{higgins2016beta, burgess2018understanding}. However, these methods make no prior or structural assumptions about what these latent variables should encode, even when some sources of variation e.g. an object's location or pose are prevalent and well-known. Recently, several methods have proposed structured models that explicitly model rotation or translation within their generative networks \cite{bepler2019explicitly, zhong2021cryodrgn, mildenhall2020nerf} by formulating the image generative model as a function mapping coordinates in space to pixel values. Although promising, only the generative portion of these methods is equivariant to rotation and translation, and the inference networks have lackluster performance due to poor inductive bias for these structured latents.

To address this, we propose TARGET-VAE, a Translation and Rotation Group Equivariant Variational Auto-Encoder. TARGET-VAE is able to learn semantic object representations that are invariant to pose and location from images corrupted by these transformations, by structurally decomposing the image generative factors into semantic, rotation, and translation components. We perform approximate inference with a group equivariant convolutional neural network \cite{cohen2016group} and specially formulated approximate posterior distribution that allows us to disentangle the latent variables into a rotationally equivariant latent rotation, translationally equivariant latent translation, and rotation and translation invariant semantic latent variables. By combining this with a spatial generator network, our framework is completely invariant to rotation and translation, unsupervised, and fully differentiable; the model is trained end-to-end using only observed images. In comprehensive experiments, we show that TARGET-VAE accurately infers the rotation and translation of objects without supervision and learns high quality object representations even when training images are heavily confounded by rotation and translation. We then show that this framework can be used to map continuous variation in 2D images of proteins collected with cryo-EM.

\section{Related Work}
In the recent years, there has been significant progress in machine learning methods for unsupervised semantic analysis of images using VAEs \cite{bepler2019explicitly, eslami2016attend, gulrajani2016pixelvae, crawford2019spatially}, flow-based methods \cite{kingma2018glow}, Generative Adversarial Networks (GAN) \cite{radford2015unsupervised, chen2016infogan, skorokhodov2021adversarial}, and capsule networks \cite{sun2021canonical}. These methods generally seek to learn a low-dimensional representation of each image in a dataset, or a distribution over this latent, by learning to reconstruct the image from the latent variable. The latent variable, then, must capture variation between images in the dataset in order for them to be accurately reconstructed. These latents can then be used as features for downstream analysis, as they capture image content. However, these representations must capture all information needed to reconstruct an image including common transformations that are not semantically meaningful. This often results in latent representations that group objects primarily by location, pose, or other nuisance variables, rather than type. One simple approach to address this is data augmentation. In this approach, additional images are created by applying a known set of nuisance transformations to the original training dataset. A network is then trained to be invariant to these transformations by penalizing the difference between low-dimensional representations of the same image with different nuisance transforms applied \cite{taylor2018improving, chen2019clusternet}. However, this approach is data inefficient and, more importantly, does not guarantee invariance to the underlying nuisance transformations. This general approach has improved with recent contrastive learning methods \cite{chen2020simple, he2020momentum, xie2021detco}, but these ultimately suffer from the same underlying problems. 


Other methods approach this problem by applying constraints on the unstructured latent space \cite{higgins2016beta, burgess2018understanding, chen2016infogan}. In this general class of methods, the latent space is not explicitly structured to disentangle semantic latent variables from latents representing other transformations. Instead, the models need to be inspected post-hoc to determine if a meaningfully separated concepts have been encoded in each latent variable and what those concepts are. A few works have addressed explicitly disentangling the latent space into structured and unstructured variables. \citet{sun2021canonical} propose a 3D point cloud representation of objects decomposed into capsules. Using a self-supervised learning scheme, their model learns transformation-invariant descriptors along with transformation-equivariant poses for each capsule, improving reconstruction and clustering. \citet{bepler2019explicitly} propose spatial-VAE, a VAE framework that divides the latent variables into unstructured, rotation, and translation components. However, only the generative part of spatial-VAE is equivariant to rotation and translation and the inference network struggles to perform meaningful inference on these transformations.

To disentangle object semantic representations from spatial transformations, the inference network should also be equivariant to the spatial transformations. Translation-equivariance can be achieved by using convolutional layers in the inference model. However, convolutional layers are not rotation-equivariant. There has been many studies in designing transformation-equivariant and specifically rotation-equivariant neural networks \cite{cohen2016group, lenssen2018group, cohen2018spherical, bekkers2019b}. In Group-equivariant Convolutional Neural Networks (G-CNNs) \cite{cohen2016group}, 2d rotation space is discretized and filters are applied over this rotation dimension in addition to the spatial dimensions of the image. This creates a network that is structurally equivariant to the discrete rotation group at the cost of additional compute. G-CNNs allow us to create a rotation-equivariant inference network.

A number of studies have proposed spatial transformation equivariant generative models, by modeling an image as a function that maps 2D or 3D spatial coordinates to the values of the pixels at those coordinates \cite{zhong2021cryodrgn, mildenhall2020nerf, stanley2007compositional, bricman2018coconet, bepler2019explicitly}. Due this mapping, any transformation of the spatial coordinates produces exactly the same transformation of the generated image. However, other than \citet{bepler2019explicitly}, none of these studies perform inference on spatial transformations. To the best of our knowledge, TARGET-VAE is the first method to have rotation and translation equivariance in both the inference and the generative networks to achieve disentangling.

\begin{figure}[t]
  \centering
    \includegraphics[width=\textwidth]{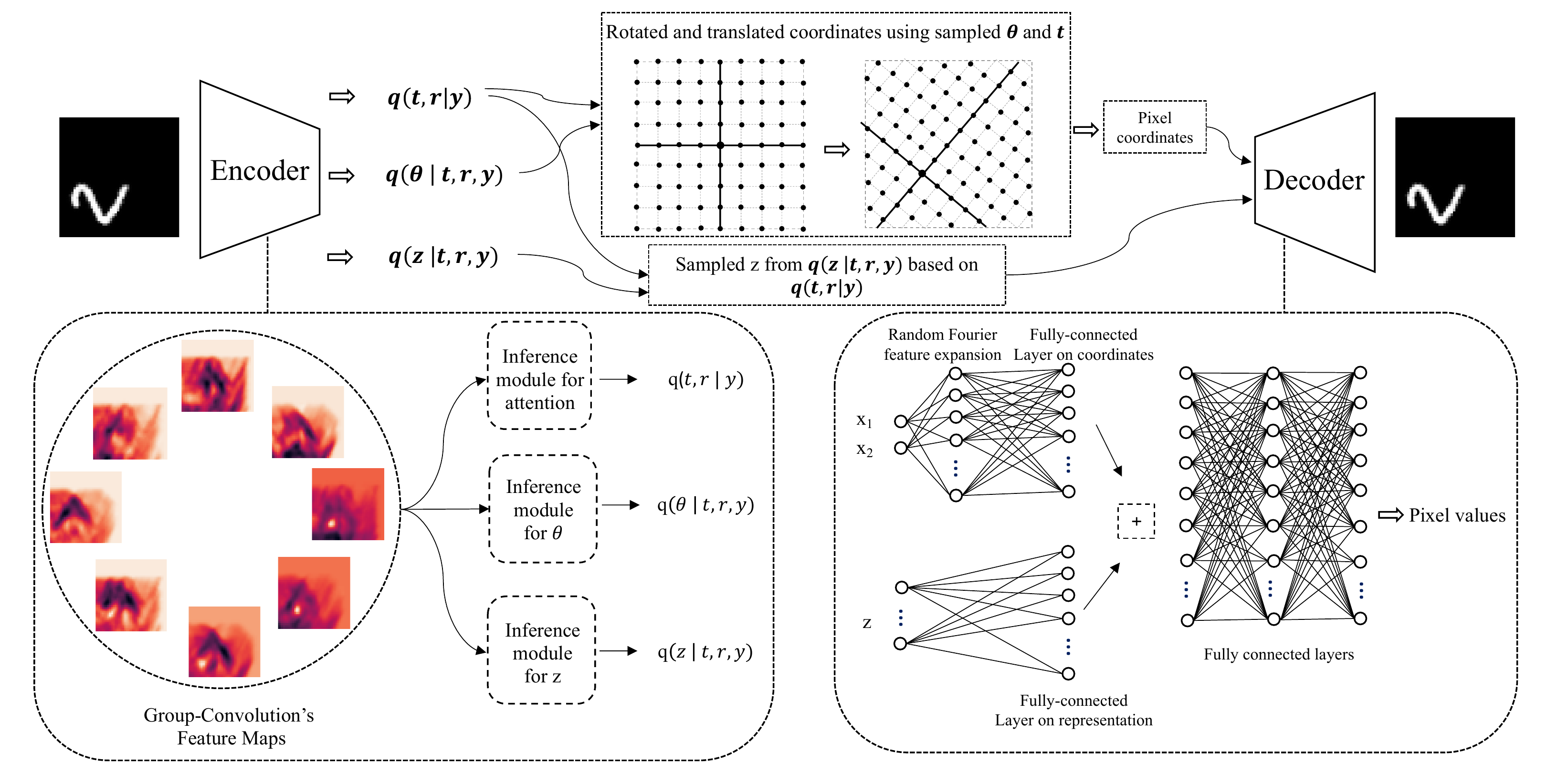}
    \caption{The TARGET-VAE framework. The encoder uses group-equivariant convolutional layers to output mixture distributions over semantic representations, rotation, and translation. The transformation-equivariant generator reconstructs the image based on the representation value $z$, and the transformed coordinates of the pixels.}
    \label{fig:model}
\end{figure}

\section{Method}

TARGET-VAE disentangles image latent variables into an unstructured semantic vector, and rotation and translation latents (Figure \ref{fig:model}). The approximate posterior distribution is factorized into translation, rotation, and content components where the joint distribution over translation and discrete rotation groups is defined by an attention map output by the equivariant encoder. The fine-grained rotation and content vectors are drawn from a mixture model where each of the joint translation and discrete rotation combinations are components of the mixture and are used to select mean and standard deviations of the continuous latents output by the encoder for each possible rotation and translation. Sampled translation, rotation, and content vectors are fed through the spatial decoder to generate images conditioned on the latent variables.


\subsection{Image generation process}
Images can be considered as the combination of discreetly identifiable pixels, and the spatial transformation of an image is equivalent to transforming the spatial coordinates of its pixels. To have an image generation process adherent to the transformations identified in the latent space, we define our generator as a function which maps the spatial coordinates of the pixels to their values. The generator outputs a probability distribution as $p(\hat{y}_{i}|z,x_{i})$, where $z$, $x_{i}$, and $\hat{y}_{i}$ are the latent content vector, the spatial coordinate of the $i$th pixel, and the value of that pixel, respectively. Similar to \cite{kingma2013auto}, for an image with $n$ pixels, we can define the probability of the image generated in this manner as sum over probability of its pixel values:
\begin{equation}
\label{eq:img_gen}
log\hspace{0.1cm}p(\hat{y}|z) = \sum_{i=1}^{n} log\hspace{0.1cm}p(\hat{y}_{i}|z,x_{i}) 
\end{equation}
To define spatial coordinates over the pixels of the image, we use Cartesian coordinates with the origin at the center of the image. Translation and rotation of the image is achieved by shifting the origin and rotating coordinates around it, respectively. Depending on the value of the content vector, $z$, the generator acts as a function over the pixel coordinates to produce an image. In this setup, the generation process is equivariant to transformations of the coordinate system. Assuming $R(\theta)$ as the rotation matrix for angle $\theta$, and translation value of $t$, the probability of the generated image from Formula \ref{eq:img_gen} is
\begin{equation}
\label{eq:img_gen_r_x}
log\hspace{0.1cm}p(\hat{y}|z, \theta, t) = \sum_{i=1}^{n} log\hspace{0.1cm}p(\hat{y}_{i}|z,R(\theta)x_{i} + t) 
\end{equation}
In this study, we focus specifically on rotation and translation of the spatial coordinates, but the image generative process extends to any transformation of the coordinate space.

\subsection{Approximate inference on latent variables}
\label{sec:inference}
We implement a VAE framework to perform approximate inference on content, rotation, and translation latent variables. Since rotation optimization is non-convex, we implement a mixture distribution over the 2D rotation space that allows the model to choose from $r$ discrete components. These components are used to approximate the posterior distributions over the rotation angle $\theta$. We represent the overall approximate posterior as $q(z,\theta,t,r|y)$, where $z$ is the latent content vector, $\theta$ is the rotation angle, $t$ and $r$ refer to the translation and discretized rotation components, and $y$ is the input image. By making the simplifying assumption that $q(z|t,r,y)$ and $q(\theta|t,r,y)$ are independent, the approximate posterior distribution factorizes as
\begin{equation}
\label{eq:factorized_posteriors}
    q(z,\theta,t,r|y) = q(z|t,r,y)q(\theta|t,r,y)q(t,r|y) \text{,}
\end{equation}

where $q(t,r|y)$ is the joint distribution over discrete translations and rotations and $q(\theta|t,r,y)$ and $q(z|t,r,y)$ are the distributions over the real-valued rotation and the latent content vector conditioned on $t$, $r$, and the input $y$.

The Kullback-Leibler (KL) divergence between the approximate posterior and the prior over the latent variables is then

\begin{align}
    \label{eq:kl_first}
    &KL(q(z,\theta,t,r|y)||p(z,\theta,t,r)) = \sum_{z,\theta,t,r}q(z,\theta,t,r|y)log\frac{q(z,\theta,t,r|y)}{p(z,\theta,t,r)}  \text{.} 
\end{align}

To simplify this equation, we use the factorization from Equation \ref{eq:factorized_posteriors} to reduce the KL-divergence to the following (see Appendix A1 for the full derivation), with the joint prior factorized into independent priors over the latent variables,

\begin{equation}
    \label{eq:kl_final}
    KL(q(z,\theta,t,r|y)||p(z,\theta,t,r)) = KL_{t,r} + \sum_{t,r} q\left(t,r\right) \left(KL_{\theta} + KL_{z} \right) \text{, where}
\end{equation}

\begin{align*}
    KL_{t,r} &= \sum_{t,r}q(t,r|y)log\frac{q(t,r|y)}{p(t,r)} \text{,} \\
    KL_{\theta} &= KL\left(q\left(\theta|t,r,y\right)||p\left(\theta|r\right) \right) \text{, and} \\
    KL_{z} &= KL\left(q\left(z|t,r,y\right)||p\left(z\right)\right) \text{.}
\end{align*}


Multiplication of $Kl_{\theta}$ and $KL_{z}$ with $q(t,r)$ in Equation \ref{eq:kl_final}, weights the KL-divergence on $\theta$ and $z$ by the joint posterior distribution over $t$ and $r$. We choose to make the priors, $p(t)$ and $p(r)$, independent giving $p(t,r) = p(t)p(r)$, but this is straightforward to relax if a non-independent prior is desired.

We define the prior on $\theta$ to be independent of the translation and $z$ to be independent of the rotation and translation. We model $\theta$ as being drawn from a mixture distribution where the mixture components are the discrete rotations, $p(\theta)=\sum_{r}p(\theta|r)p(r)$. Given $r$, $\theta$ is drawn from a Gaussian distribution with mean given by the angle offset defined by $r$ and a fixed standard deviation which depends on the number of discrete rotation groups and is set to $\frac{\pi}{r}$. We tighten each of the $\theta$ mixture distributions by shrinking the standard deviation as $r$ grows to reduce overlap between these distributions when the number of discrete rotations is large. For the prior on $r$, we usually use a uniform distribution, but a discretized normal distribution can also be used to bias the model towards certain rotation angles if a non-uniform rotation distribution is known a priori. Then, $p(r)$ is calculated based on $\theta_{offset}$ for each discrete rotation. For example, when $r=4$, then $\theta_{offset} \in \{0, \frac{\pi}{2}, \pi,\frac{3\pi}{2}\}$, which are the rotation angles of the discrete rotation group, P$_4$. The prior over translation, $t$, is Gaussian with mean 0, which denotes the center of the 2D coordinate system. The standard deviation of this prior depends on the number of pixels in our fixed-size coordinate system and whether we want the model to be flexible with large translation values or not. Lower standard deviation for the prior over translations, will more heavily penalize large predicted translation values.

The full variational lower-bound for our model is
\begin{equation}
    \label{eq:elbo}
    \underset{(z, \theta, t)\sim\ q(z, \theta, t|y)}
    {\mathbb{E}}[log\hspace{0.1cm}p(\hat{y}|z, \theta, t)] \hspace{0.2cm} - \hspace{0.2cm} KL(q(z,\theta,t,r|y)||p(z,\theta,r,t)) \text{,}
\end{equation}

where the KL-divergence term is defined in Equation \ref{eq:kl_final}. 

\subsection{Neural network architecture}
\label{subsec:arch}

\subsubsection{Group convolutional encoder}
Convolutional layers are translation equivariant, meaning that a spatial shift in the image causes the same shift in the output feature map, but are not rotation-equivariant. To incorporate rotation-equivariance in the inference model, we use group convolutional layers \cite{cohen2016group}. A symmetry group $G$ is defined as a set $X$ with an operation $.$, which is associative on the set, and has inverse and identity elements. Furthermore, combination of two symmetry transformations on a set, also creates a symmetry group. 

We define the P$_{r}$ group where $r \in \mathbb{N}$, and the operations consist of all translations, and $r$ discrete rotations, about any center of rotation in a 2D space. In the group convolutional layer, each kernel is rotated by $k\frac{2\pi}{r}$ angles, where $k \in \{0,1, ..., r-1\}$ and is then convolved with the input to produce an output map with $r$ values corresponding to the rotations of the kernels.

Following the group convolutional layer, we apply three 1x1 group convolutional layers (Figure \ref{fig:model}), which are efficiently implemented as 3D convolutions. The final layer outputs the parameters of the approximate posterior distributions, $q(t=(i,j),r=r'|y) = \text{softmax}(a(y))_{i,j,r'}$ where $a(y)$ is an $\text{r} x \text{N} x \text{M}$ activation map output by the network given input $y$, and $q(z | t=(i,j), r=r', y) = \mathcal{N}(\mu_z(y)_{i,j,r'}, \sigma_z(y)_{i,j,r'}^2)$ and $q(\theta | t=(i,j), r=r', y) = \mathcal{N}(\mu_\theta(y)_{i,j,r'} + \frac{r'2\pi}{r}, \sigma_\theta(y)_{i,j,r'}^2)$ where $\mu_z(y)_{i,j,r'}$, $\sigma_z(y)_{i,j,r'}^2$, $\mu_\theta(y)_{i,j,r'}$, and $\sigma_\theta(y)_{i,j,r'}^2$ are output by the inference network for each spatial location and discrete rotation group. $\frac{r'2\pi}{r}$ is the angle of the $r'$th discrete rotation.


During training, we sample from the joint posterior distribution of $q(t, r | y)$ using Gumbel-Softmax \cite{jang2016categorical} in order to differentiably sample from the approximate posterior to estimate the expected reconstruction error given $q$. We use the soft one-hot sample to calculate the weighted sum of the parameters of the $q(z | t,r,y)$ and $q(\theta | t,r,y)$ distributions and then sample from those distributions to get a sample from the full joint approximate posterior. We define a coordinate system where $x$ is in the [-1,1] range, and is centered on the middle of the image. The output of the Gumbel-Softmax directly describes a sample from the approximate posterior over the translation, which we get by calculating the weighted sum over the translation grid points.

\subsubsection{Spatial decoder}

Given $t$ and $\theta$, we translate and rotate the image spatial coordinates, $x' = R(\theta)x + t$, where $R(\theta)$ is the rotation matrix for angle $\theta$. The generator receives the sampled $z$ value along with transformed spatial coordinates, $x'$, and outputs the value for the pixels at those spatial coordinates, forming the image. We use random Fourier feature expansion \cite{rahimi2007random} on the coordinates of the pixel before feeding it through a stack of fully-connected layers (Figure \ref{fig:model}). The spatial coordinate and $z$ inputs are processed separately by two parallel sets of fully-connected layers. Those representations are then summed and processed by the remaining shared fully-connected layers which give the final output value.


\section{Results}
\subsection{Experiment setup}
We run our experiments with P$_{4}$, P$_{8}$, and P$_{16}$ group convolutional layers in the inference model. The input images are normalized to the range of [0,1]. We use 128 rectangular kernels where size of the kernels in the first layer, is set to a value larger than the size of the targeted objects. The reason for this is that we only use one group convolutional layer in our models and we want to assure that its receptive field is at least as large as the objects of interest. We use leaky-ReLU activation functions, and the batch size for all the experiments is set to 100. We use ADAM optimizer \cite{kingma2014adam}, with learning rate of 2e-4, and the learning rate is decayed by the factor of 0.5 after no improvements in the loss for 10 epochs. We run the training for a maximum of 500 epochs with early-stopping in case of no improvements in the loss for 20 epochs. We run all our experiments in a single NVIDIA A100 GPU, with 80 GB memory.

Each fully-connected layer in the generator has 512 hidden units and the generator output is a probability value, in the case of the binary valued pixels, or mean and standard deviation for real valued pixels. For RGB images, the generator outputs R, G, and B values for each spatial location.

\subsection{TARGET-VAE predicts the translation and rotation values of objects with high correlation}
\label{subsec:results_corr}

We first examine whether TARGET-VAE can accurately predict rotation and translation of objects despite being trained without supervision.

\paragraph{Datasets} We use two variants of the MNIST dataset: 1) (MNIST(N)) MNIST digits randomly rotated by sampling from $\mathcal{N}(0, \frac{\pi^{2}}{16})$ and randomly translated by sampling from $\mathcal{N}(0, 5^{2})$ pixels, and 2) (MNIST(U)) MNIST digits rotated by sampling uniformly from \textit{U}$(0, 2\pi)$ and translated using the same distribution as in 1) (Appendix Figure \ref{fig:app_mnist_n_u}). In both datasets, the train and test sets have 60,000, and 10,000 images of dimensions 50x50 pixels, respectively.


\paragraph{Training} We train TARGET-VAE with P$_{4}$, P$_{8}$, and P$_{16}$ and z\_dim=2 on these generated datasets. For the prior over $\theta$, we use $\mathcal{N}(0,\frac{\pi}{4})$ for the MNIST(N) dataset, and \textit{U}$(0, 2\pi)$ for the MNIST(U). For $p(\theta|r)$ we calculate $p(\theta_{offset}) \sim\ p(\theta)$, where $\theta_{offset}$ are the rotation angles of the kernels in each dimension of r. For both datasets, we set the prior over $t$ to $\mathcal{N}(0,5^{2})$. 

Given the trained model, we perform inference on $t$ and $\theta$ on the test set using the inference network. We find the most likely values of $t$ and $\theta$ from the approximate posterior distributions given by the network and compare these with the ground truth values for the test set images. We calculate the Pearson correlation between the predicted and true translations and the circular correlation between our predicted and true rotation angles \cite{jammalamadaka2001topics}. In a comparison with a spatial-VAE network \cite{bepler2019explicitly} trained with the same number of parameters as TARGET-VAE and z\_dim=2, we find that TARGET-VAE significantly improves over spatial-VAE when predicting rotation. Spatial-VAE fails to give meaningful rotation predictions when digits are uniformly rotated, but TARGET-VAE predicts rotation with remarkable accuracy (Table \ref{tab:mnist_corr}). We also find that increasing the number of discrete rotations in the group convolution improves the accuracy of rotation inference, but does not have a significant impact on translation inference. Interestingly, spatial-VAE performs better than TARGET-VAE on translation inference, though the difference is small and predictions from both correlate strongly with the ground truth. One possible reason for this is that TARGET-VAE only predicts whole pixel translations, due to the formulation of the approximate posterior, whereas spatial-VAE predicts real valued translations and, therefore, can predict sub-pixel values. In the future, including fine-grained translation inference in TARGET-VAE, similarly to the rotation distribution, could allow TARGET-VAE to resolve sub-pixel translations as well. We also found that increasing the dimension of $z$ did not have a noticeable effect on the ability of the network to predict translation and rotation.

\begin{table}[ht]
  \caption{Translation and rotation correlation on MNIST(N) and MNIST(U) with z\_dim=2}
  \label{tab:mnist_corr}
  \centering
  \begin{tabular}{llllll}
    \toprule
    & \multicolumn{2}{c}{MNIST(N)}   & \multicolumn{3}{c}{MNIST(U)}    \\
    \cmidrule(r){2-3} \cmidrule(r){5-6}
    Group Convolution     & Translation     & Rotation &  & Translation     & Rotation\\
    \midrule
    Spatial-VAE \cite{bepler2019explicitly}  & \textbf{0.994, 0.993}    & 0.564  &   & \textbf{0.982, 0.983}    & 0.005  \\
    TARGET-VAE P$_{4}$  & 0.97, 0.972    & 0.814  &   & 0.975, 0.976    & 0.80  \\
    
    TARGET-VAE P$_{8}$  & 0.981, 0.981  & 0.89  &   & 0.972, 0.971  & 0.859   \\
    
    TARGET-VAE P$_{16}$  & 0.969, 0.972   & \textbf{0.898}  &   & 0.974, 0.971  & \textbf{0.93}   \\
    \bottomrule
  \end{tabular}
\end{table}

\begin{figure}[ht]
  \centering
    \includegraphics[width=\textwidth]{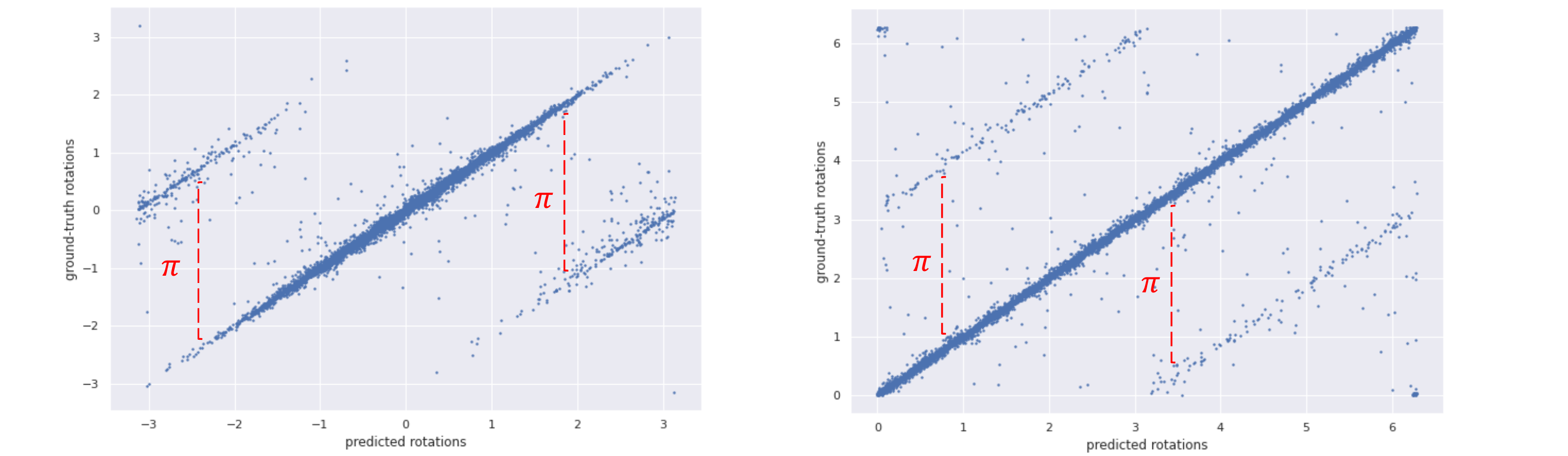}
    \caption{Left: Predicted $\theta$ vs. ground-truth data in MNIST(N), Right: Predicted $\theta$ vs. ground-truth data in MNIST(U), with TARGET-VAE P$_{8}$. There is a $\pi$ difference between the predicted angles and the ground-truth values of some of the digits (digits 0, 1, and 8), which is related to the rotation symmetry of them. }
    \label{fig:mnist_pi_diff}
\end{figure}

Some of the digits in MNIST have natural ambiguity in their rotations. Investigating the correlation between the predicted rotation values and the ground truth ones shows a difference about $\pi$ between some of the predicted angles and the ground truth (Figure \ref{fig:mnist_pi_diff}). This is expected, because several digits have approximate symmetry. For example, 0, 1, and 8 are roughly 2-fold symmetric (Appendix A.3) creating inherent ambiguity in the rotation that is reflected in larger error in the rotation predictions (Table \ref{tab:rotError}). Six and 9 are also similar when rotated by $\pi$, although, interestingly, TARGET-VAE is able to distinguish these digits and predict their rotations accurately. Perhaps this is because 9 is typical drawn with a straight spine while 6 is drawn with a curved one. This also suggests that the posterior predictions given by TARGET-VAE could be helpful for identifying the symmetry group of objects.

\subsection{TARGET-VAE improves clustering accuracy by learning translation and rotation invariant features}
\label{subsec:results_clustering}

By disentangling rotation and translation from the representation values, TARGET-VAE is able to recover semantic organization of objects. Using the z\_dim=2 models trained above and additional TARGET-VAE models trained with z\_dim=32, we extract the most likely value of $z$ for each model from $q(z|t,r,y)$ for each image in the MNIST(N) and MNIST(U) test sets. To evaluate the disentangled semantic value of these representations, we cluster the test set images in $z$-space using agglomerative clustering \cite{murtagh2014ward}. We then calculate the clustering accuracy by finding the highest accuracy assignment between clusters and digits for each set of representations. 
We find that semantic representations produced by TARGET-VAE much more closely represent the underlying semantics of the MNIST dataset when compared to representations produced by baseline methods (Table \ref{tab:mnist_clustering}). TARGET-VAE achieves clustering accuracies approximately 2-5x higher than standard VAE \cite{kingma2013auto}, beta-VAE \cite{higgins2016beta}, and spatial-VAE \cite{bepler2019explicitly} models trained with the same number of parameters as TARGET-VAE. We set the z\_dim=2 for both spatial-VAE and TARGET-VAE. Since the standard VAE and the beta-VAE do not have dedicated latent variables for identifying translation and rotation, we set the dimension of the unstructured latent space to 4 for these models. Also, we use $\beta=4$ for the beta-VAE based on the settings proposed by \citet{higgins2016beta}. We believe that direct inference on rotation and translation with the equivariant inference network of TARGET-VAE plus the transformation-equivariant generator are the main reason for the dramatic increase in the clustering accuracy (see ablation studies in Appendix A.7).

\begin{figure}[ht]
  \centering
    \includegraphics[width=0.9\textwidth]{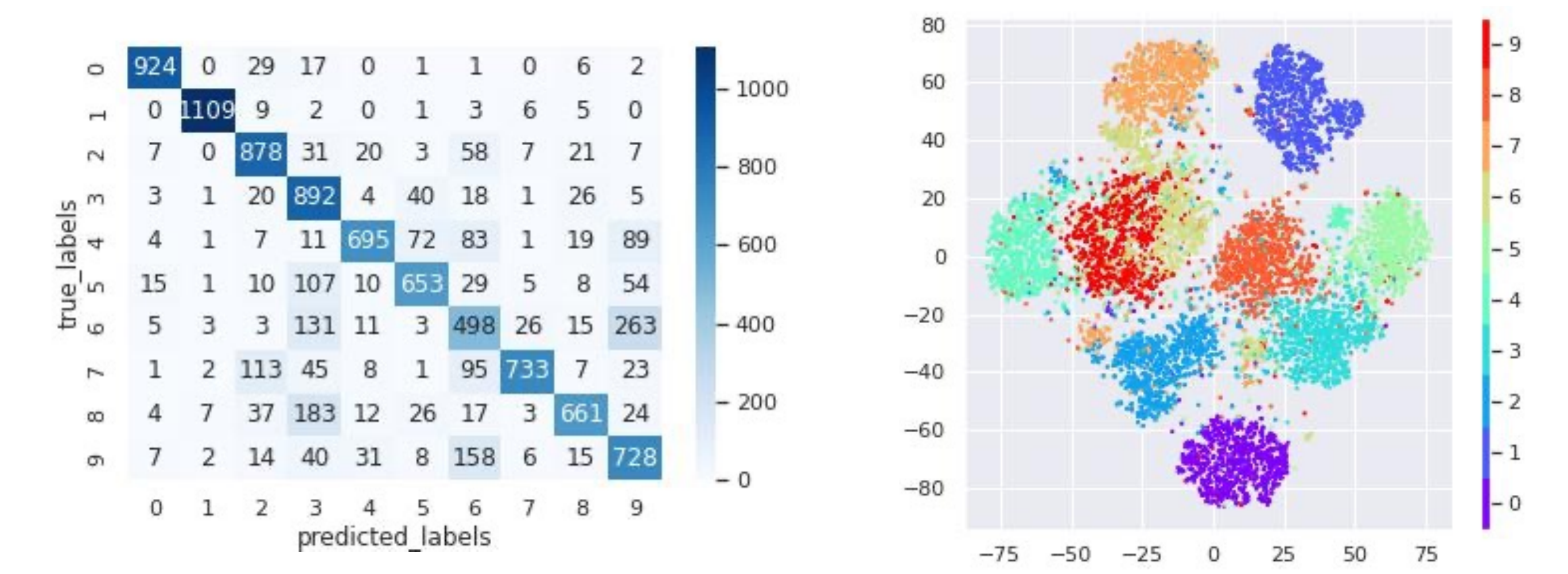}
    \caption{Confusion matrix of trained TARGET-VAE on MNIST(N) with z\_dim=32 (left), and the t-sne figure of the 32 dimension latent representation space (right). Based on the confusion matrix, most of the digits are clustered correctly, and the largest misclassification is related to the confusion between rotated digits 6 and 9, which is understandable.}
    \label{fig:p8_tsne_conf}
\end{figure}

We also observe that increasing the dimension of $z$ can significantly increase clustering performance and produces clearly distinct clusters for each digit (Figure \ref{fig:p8_tsne_conf}). Increasing the number of discrete rotations also can improve clustering performance, likely because improving rotation inference means that $z$ and $\theta$ are better disentangled, but this trend is not universally true for both MNIST(N) and MNIST(U). We note that the P$_8$ model with z\_dim=32 performs best on MNIST(N), perhaps because rotation inference is performed roughly equally well by the P$_8$ and P$_{16}$ models on this dataset.

\begin{table}
  \caption{Clustering accuracy (\%) on MNIST(N) and MNIST(U)}
  \label{tab:mnist_clustering}
  \centering
  \begin{tabular}{lll}
    \toprule
    Model  & MNIST(N)   &MNIST(U)   \\
    \midrule
    VAE (z\_dim=4) \cite{kingma2013auto}  & 15.3    & 12.8  \\
    Beta-VAE (z\_dim=4) \cite{higgins2016beta}  & 15.1    & 18.0  \\
    Spatial-VAE (z\_dim=2) \cite{bepler2019explicitly}  & 37.1    & 28.2  \\
    \midrule
    TARGET-VAE P$_{4}$ (z\_dim=2)  & 56.4   & 56.6    \\
    TARGET-VAE P$_{8}$ (z\_dim=2)  & 60.1   &  57.1    \\
    TARGET-VAE P$_{16}$ (z\_dim=2)  & 60.1   & 63.4   \\
    \midrule
    TARGET-VAE P$_{4}$ (z\_dim=32)  & 65.1   & 64.3    \\
    TARGET-VAE P$_{8}$ (z\_dim=32)  & \textbf{77.7}   &  69.1    \\
    TARGET-VAE P$_{16}$ (z\_dim=32)  & 75.2   & \textbf{71.2}  \\
    \bottomrule
  \end{tabular}
\end{table}

TARGET-VAE is also able to detect multiple objects at inference time (Appendix A.4) and is able to learn meaningfully disentangled semantic representations on other datasets, including dSprites \cite{dsprites17} (Appendix A.5) and the galaxy zoo \cite{lintott2008galaxy} (Appendix A.6).

\subsection{Identifying class-averages and protein conformations in the cryo-EM micrographs with TARGET-VAE}
In single-particle cryo-EM micrographs, individual biomolecules (particles) are spread throughout images with random orientations and locations. Being 2d projections of 3d structures, particle images have variability due to variation in the orientation of the particles in sample and also have variability due to conformational heterogeneity between structures. To be able to identify the particles and their conformations, semantic representations should be invariant to these transformations. Here, we demonstrate the capacity of TARGET-VAE to learning different classes and conformations of particles in cryo-EM micrographs.

\paragraph{Identifying different classes of particles in EMPIAR-10025.} 

\begin{figure}[ht]
  \centering
    \includegraphics[width=\textwidth]{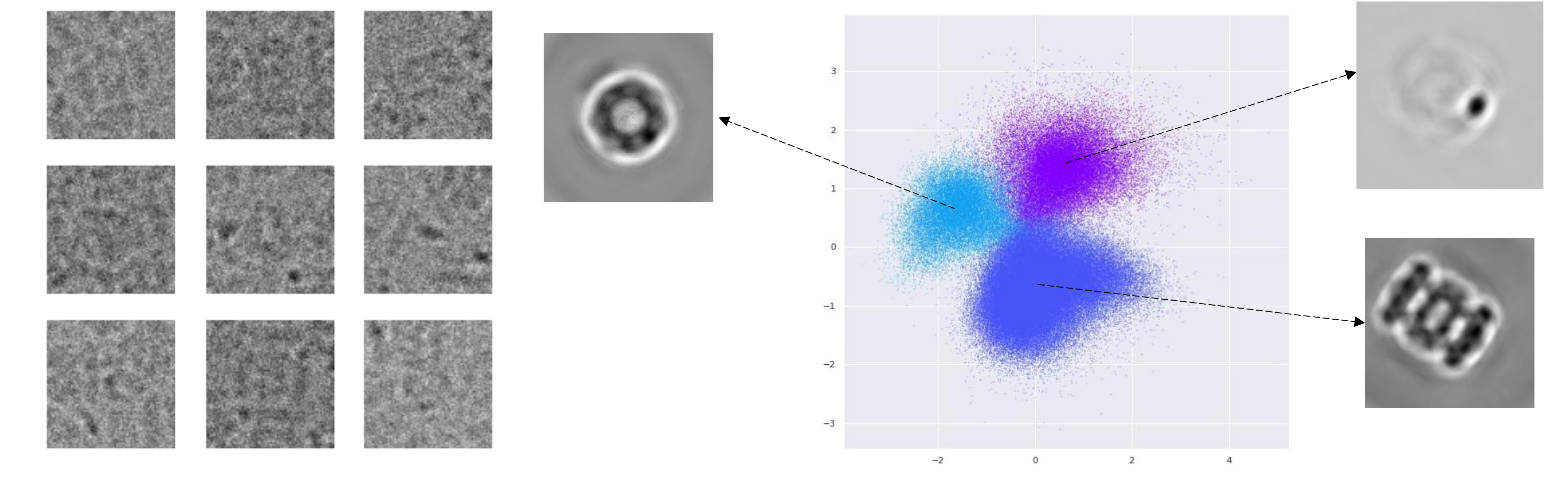}
    \caption{Left: Images of T20S proteasome particles from EMPIAR-10025. These images are extremely noisy. Right: The 2D latent representation space for EMPIAR-10025 learned by TARGET-VAE. TARGET-VAE identifies three clusters from which we show reconstructed examples produced by the spatial generator network.}
    \label{fig:latent_10025}
\end{figure}

The dataset EMPIAR-10025 \cite{campbell20152} contains T20S proteasome micrographs used to solve the T20S proteasome structure at 2.8 {\AA} resolution. We use the Topaz \cite{bepler2019positive} to create a stack of 161,292 400x400 images of T20S proteasome particles. After downsampling by factor of 4 and normalizing the images, we train a P$_8$ TARGET-VAE with z\_dim=2. Due to the size and variability of these images, we use a larger spatial generator network than the one used on the MNIST datasets. Here, the generator has six fully-connected layers where each layer has 512 hidden units. We use a uniform prior over $r$. We randomly select 10\% of the images as the validation set to monitor the training process. After training, we apply the inference model on the images of the particles to extract the most-probable semantic representations (same as the clustering of the MNIST datasets described in \ref{subsec:results_clustering}). We find that these representations produce meaningful semantic clustering of the particles, grouping them by view and the presence of contaminants (gold particles in the dataset), and also can be used to reconstruct the particles with specified rotation and translation (Figure \ref{fig:latent_10025}).

\paragraph{Identifying different views of particles in EMPIAR-10029.}

\begin{figure}[ht]
  \centering
    \includegraphics[width=\textwidth]{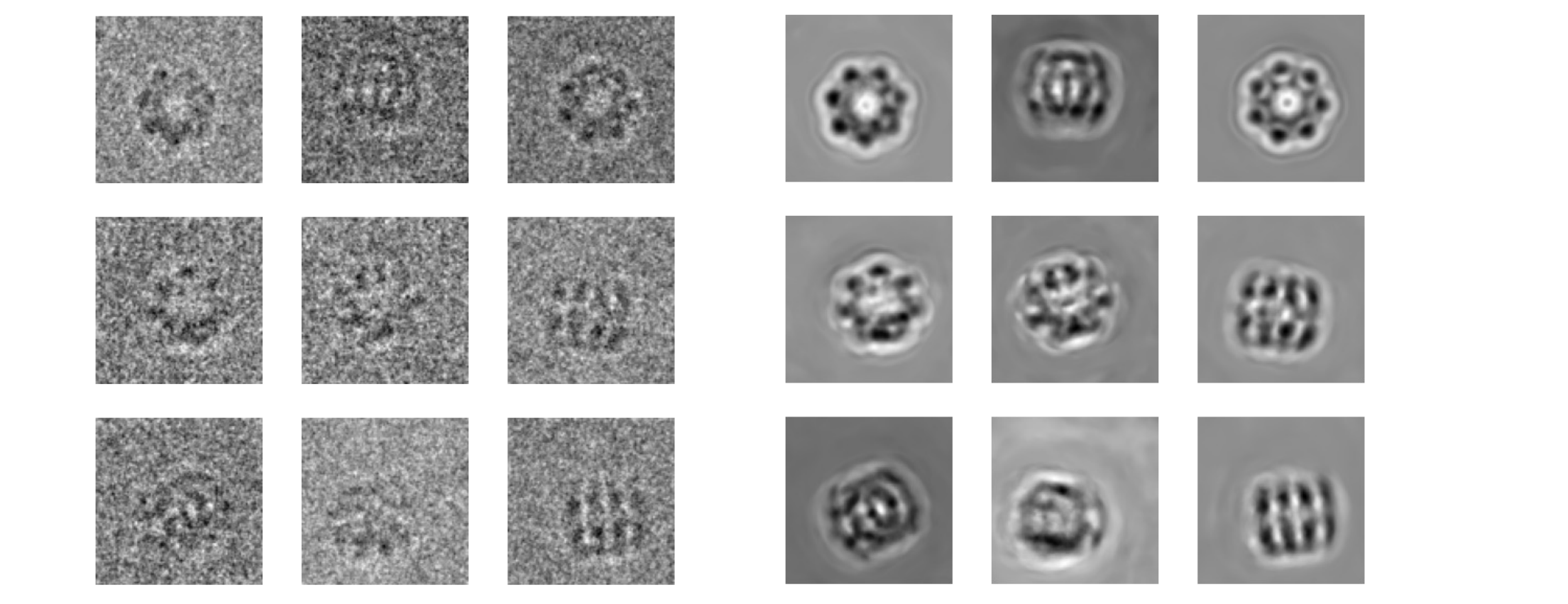}
    \caption{Left: Simulated GroEL particle images from EMPIAR-10029. Right: The reconstructed particles. TARGET-VAE learns the distribution over particle views controlled for in plane rotation and translation and generates de-noised particle views.}
    \label{fig:empiar_10029}
\end{figure}

EMPIAR-10029 is a simulated EM dataset of GroEL particles. This dataset has 10,000 200x200 images, which we downsample to 100x100 pixels and normalize. We train the TARGET-VAE with the same settings as for EMPIAR-10025. The latent space does not identify separate clusters in this dataset, likely because a continuum over views is present from simulation, but exploration of the learned semantic latent space identifies GroEL projections independent of their rotation and translation (Figure \ref{fig:empiar_10029}).

TARGET-VAE is also able to discover continuous latent heterogeneity in other cryo-EM datasets (Appendix A.8).

\section{Conclusion}
We present TARGET-VAE, a translation and rotation group equivariant variational autoencoder framework for learning translation and rotation invariant object representation in images without supervision. By structuring the encoder to be translation and rotation equivariant, the model learns latent rotation and translation variables, disentangling these transformations from object semantics encoded in a separately learned unstructured semantic latent variables, by training it jointly with a spatially equivariant generator network. TARGET-VAE learns to accurately predict object locations and rotations, without any supervision, and also learns content representations that reflect known semantics (i.e., clusters match known semantic labels) across multiple datasets. Although we only consider 2d images containing single objects with rotation and translation, this framework can be extended to other object transformations by adopting encoder networks with the proper equivariances and adapting the approximate posterior distribution. In the future, we expect this framework can also be extended to multi-object detection, object tracking over time, and to 3d environments. TARGET-VAE lays the groundwork for a new generation of fully unsupervised object detection and semantic analysis methods for cryo-EM and other imaging modalities.

\begin{ack}
This work was supported by a grant from the Simons Foundation (SF349247).
\end{ack}

\bibliographystyle{unsrtnat}
\bibliography{references}

\begin{thebibliography}{36}
\providecommand{\natexlab}[1]{#1}
\providecommand{\url}[1]{\texttt{#1}}
\expandafter\ifx\csname urlstyle\endcsname\relax
  \providecommand{\doi}[1]{doi: #1}\else
  \providecommand{\doi}{doi: \begingroup \urlstyle{rm}\Url}\fi

\bibitem[Lintott et~al.(2008)Lintott, Schawinski, Slosar, Land, Bamford,
  Thomas, Raddick, Nichol, Szalay, Andreescu, et~al.]{lintott2008galaxy}
Chris~J Lintott, Kevin Schawinski, An{\v{z}}e Slosar, Kate Land, Steven
  Bamford, Daniel Thomas, M~Jordan Raddick, Robert~C Nichol, Alex Szalay, Dan
  Andreescu, et~al.
\newblock Galaxy zoo: morphologies derived from visual inspection of galaxies
  from the sloan digital sky survey.
\newblock \emph{Monthly Notices of the Royal Astronomical Society},
  389\penalty0 (3):\penalty0 1179--1189, 2008.

\bibitem[Sigworth(2016)]{sigworth2016principles}
Fred~J Sigworth.
\newblock Principles of cryo-em single-particle image processing.
\newblock \emph{Microscopy}, 65\penalty0 (1):\penalty0 57--67, 2016.

\bibitem[Kingma and Welling(2013)]{kingma2013auto}
Diederik~P Kingma and Max Welling.
\newblock Auto-encoding variational bayes.
\newblock \emph{arXiv preprint arXiv:1312.6114}, 2013.

\bibitem[Higgins et~al.(2016)Higgins, Matthey, Pal, Burgess, Glorot, Botvinick,
  Mohamed, and Lerchner]{higgins2016beta}
Irina Higgins, Loic Matthey, Arka Pal, Christopher Burgess, Xavier Glorot,
  Matthew Botvinick, Shakir Mohamed, and Alexander Lerchner.
\newblock beta-vae: Learning basic visual concepts with a constrained
  variational framework.
\newblock 2016.

\bibitem[Burgess et~al.(2018)Burgess, Higgins, Pal, Matthey, Watters,
  Desjardins, and Lerchner]{burgess2018understanding}
Christopher~P Burgess, Irina Higgins, Arka Pal, Loic Matthey, Nick Watters,
  Guillaume Desjardins, and Alexander Lerchner.
\newblock Understanding disentangling in $\beta$-\uppercase{VAE}.
\newblock \emph{arXiv preprint arXiv:1804.03599}, 2018.

\bibitem[Bepler et~al.(2019{\natexlab{a}})Bepler, Zhong, Kelley, Brignole, and
  Berger]{bepler2019explicitly}
Tristan Bepler, Ellen Zhong, Kotaro Kelley, Edward Brignole, and Bonnie Berger.
\newblock Explicitly disentangling image content from translation and rotation
  with spatial-\uppercase{VAE}.
\newblock \emph{Advances in Neural Information Processing Systems}, 32,
  2019{\natexlab{a}}.

\bibitem[Zhong et~al.(2021)Zhong, Bepler, Berger, and Davis]{zhong2021cryodrgn}
Ellen~D Zhong, Tristan Bepler, Bonnie Berger, and Joseph~H Davis.
\newblock Cryodrgn: reconstruction of heterogeneous cryo-em structures using
  neural networks.
\newblock \emph{Nature methods}, 18\penalty0 (2):\penalty0 176--185, 2021.

\bibitem[Mildenhall et~al.(2020)Mildenhall, Srinivasan, Tancik, Barron,
  Ramamoorthi, and Ng]{mildenhall2020nerf}
Ben Mildenhall, Pratul~P Srinivasan, Matthew Tancik, Jonathan~T Barron, Ravi
  Ramamoorthi, and Ren Ng.
\newblock Nerf: Representing scenes as neural radiance fields for view
  synthesis.
\newblock In \emph{European conference on computer vision}, pages 405--421.
  Springer, 2020.

\bibitem[Cohen and Welling(2016)]{cohen2016group}
Taco Cohen and Max Welling.
\newblock Group equivariant convolutional networks.
\newblock In \emph{International conference on machine learning}, pages
  2990--2999. PMLR, 2016.

\bibitem[Eslami et~al.(2016)Eslami, Heess, Weber, Tassa, Szepesvari, Hinton,
  et~al.]{eslami2016attend}
SM~Eslami, Nicolas Heess, Theophane Weber, Yuval Tassa, David Szepesvari,
  Geoffrey~E Hinton, et~al.
\newblock Attend, infer, repeat: Fast scene understanding with generative
  models.
\newblock \emph{Advances in Neural Information Processing Systems}, 29, 2016.

\bibitem[Gulrajani et~al.(2016)Gulrajani, Kumar, Ahmed, Taiga, Visin, Vazquez,
  and Courville]{gulrajani2016pixelvae}
Ishaan Gulrajani, Kundan Kumar, Faruk Ahmed, Adrien~Ali Taiga, Francesco Visin,
  David Vazquez, and Aaron Courville.
\newblock Pixelvae: A latent variable model for natural images.
\newblock \emph{arXiv preprint arXiv:1611.05013}, 2016.

\bibitem[Crawford and Pineau(2019)]{crawford2019spatially}
Eric Crawford and Joelle Pineau.
\newblock Spatially invariant unsupervised object detection with convolutional
  neural networks.
\newblock In \emph{Proceedings of the AAAI Conference on Artificial
  Intelligence}, volume~33, pages 3412--3420, 2019.

\bibitem[Kingma and Dhariwal(2018)]{kingma2018glow}
Durk~P Kingma and Prafulla Dhariwal.
\newblock Glow: Generative flow with invertible 1x1 convolutions.
\newblock \emph{Advances in neural information processing systems}, 31, 2018.

\bibitem[Radford et~al.(2015)Radford, Metz, and
  Chintala]{radford2015unsupervised}
Alec Radford, Luke Metz, and Soumith Chintala.
\newblock Unsupervised representation learning with deep convolutional
  generative adversarial networks.
\newblock \emph{arXiv preprint arXiv:1511.06434}, 2015.

\bibitem[Chen et~al.(2016)Chen, Duan, Houthooft, Schulman, Sutskever, and
  Abbeel]{chen2016infogan}
Xi~Chen, Yan Duan, Rein Houthooft, John Schulman, Ilya Sutskever, and Pieter
  Abbeel.
\newblock Infogan: Interpretable representation learning by information
  maximizing generative adversarial nets.
\newblock \emph{Advances in neural information processing systems}, 29, 2016.

\bibitem[Skorokhodov et~al.(2021)Skorokhodov, Ignatyev, and
  Elhoseiny]{skorokhodov2021adversarial}
Ivan Skorokhodov, Savva Ignatyev, and Mohamed Elhoseiny.
\newblock Adversarial generation of continuous images.
\newblock In \emph{Proceedings of the IEEE/CVF Conference on Computer Vision
  and Pattern Recognition}, pages 10753--10764, 2021.

\bibitem[Sun et~al.(2021)Sun, Tagliasacchi, Deng, Sabour, Yazdani, Hinton, and
  Yi]{sun2021canonical}
Weiwei Sun, Andrea Tagliasacchi, Boyang Deng, Sara Sabour, Soroosh Yazdani,
  Geoffrey~E Hinton, and Kwang~Moo Yi.
\newblock Canonical capsules: Self-supervised capsules in canonical pose.
\newblock \emph{Advances in Neural Information Processing Systems},
  34:\penalty0 24993--25005, 2021.

\bibitem[Taylor and Nitschke(2018)]{taylor2018improving}
Luke Taylor and Geoff Nitschke.
\newblock Improving deep learning with generic data augmentation.
\newblock In \emph{2018 IEEE Symposium Series on Computational Intelligence
  (SSCI)}, pages 1542--1547. IEEE, 2018.

\bibitem[Chen et~al.(2019)Chen, Li, Xu, Chen, Wang, and
  Lin]{chen2019clusternet}
Chao Chen, Guanbin Li, Ruijia Xu, Tianshui Chen, Meng Wang, and Liang Lin.
\newblock Clusternet: Deep hierarchical cluster network with rigorously
  rotation-invariant representation for point cloud analysis.
\newblock In \emph{Proceedings of the IEEE/CVF conference on computer vision
  and pattern recognition}, pages 4994--5002, 2019.

\bibitem[Chen et~al.(2020)Chen, Kornblith, Norouzi, and Hinton]{chen2020simple}
Ting Chen, Simon Kornblith, Mohammad Norouzi, and Geoffrey Hinton.
\newblock A simple framework for contrastive learning of visual
  representations.
\newblock In \emph{International conference on machine learning}, pages
  1597--1607. PMLR, 2020.

\bibitem[He et~al.(2020)He, Fan, Wu, Xie, and Girshick]{he2020momentum}
Kaiming He, Haoqi Fan, Yuxin Wu, Saining Xie, and Ross Girshick.
\newblock Momentum contrast for unsupervised visual representation learning.
\newblock In \emph{Proceedings of the IEEE/CVF conference on computer vision
  and pattern recognition}, pages 9729--9738, 2020.

\bibitem[Xie et~al.(2021)Xie, Ding, Wang, Zhan, Xu, Sun, Li, and
  Luo]{xie2021detco}
Enze Xie, Jian Ding, Wenhai Wang, Xiaohang Zhan, Hang Xu, Peize Sun, Zhenguo
  Li, and Ping Luo.
\newblock Detco: Unsupervised contrastive learning for object detection.
\newblock In \emph{Proceedings of the IEEE/CVF International Conference on
  Computer Vision}, pages 8392--8401, 2021.

\bibitem[Lenssen et~al.(2018)Lenssen, Fey, and Libuschewski]{lenssen2018group}
Jan~Eric Lenssen, Matthias Fey, and Pascal Libuschewski.
\newblock Group equivariant capsule networks.
\newblock \emph{Advances in Neural Information Processing Systems}, 31, 2018.

\bibitem[Cohen et~al.(2018)Cohen, Geiger, K{\"o}hler, and
  Welling]{cohen2018spherical}
Taco~S Cohen, Mario Geiger, Jonas K{\"o}hler, and Max Welling.
\newblock Spherical cnns.
\newblock \emph{arXiv preprint arXiv:1801.10130}, 2018.

\bibitem[Bekkers(2019)]{bekkers2019b}
Erik~J Bekkers.
\newblock B-spline cnns on lie groups.
\newblock In \emph{International Conference on Learning Representations}, 2019.

\bibitem[Stanley(2007)]{stanley2007compositional}
Kenneth~O Stanley.
\newblock Compositional pattern producing networks: A novel abstraction of
  development.
\newblock \emph{Genetic programming and evolvable machines}, 8\penalty0
  (2):\penalty0 131--162, 2007.

\bibitem[Bricman and Ionescu(2018)]{bricman2018coconet}
Paul~Andrei Bricman and Radu~Tudor Ionescu.
\newblock Coconet: A deep neural network for mapping pixel coordinates to color
  values.
\newblock In \emph{International Conference on Neural Information Processing},
  pages 64--76. Springer, 2018.

\bibitem[Jang et~al.(2016)Jang, Gu, and Poole]{jang2016categorical}
Eric Jang, Shixiang Gu, and Ben Poole.
\newblock Categorical reparameterization with gumbel-softmax.
\newblock \emph{arXiv preprint arXiv:1611.01144}, 2016.

\bibitem[Rahimi and Recht(2007)]{rahimi2007random}
Ali Rahimi and Benjamin Recht.
\newblock Random features for large-scale kernel machines.
\newblock \emph{Advances in neural information processing systems}, 20, 2007.

\bibitem[Kingma and Ba(2014)]{kingma2014adam}
Diederik~P Kingma and Jimmy Ba.
\newblock Adam: A method for stochastic optimization.
\newblock \emph{arXiv preprint arXiv:1412.6980}, 2014.

\bibitem[Jammalamadaka and Sengupta(2001)]{jammalamadaka2001topics}
S~Rao Jammalamadaka and Ambar Sengupta.
\newblock \emph{Topics in circular statistics}, volume~5.
\newblock world scientific, 2001.

\bibitem[Murtagh and Legendre(2014)]{murtagh2014ward}
Fionn Murtagh and Pierre Legendre.
\newblock Ward’s hierarchical agglomerative clustering method: which
  algorithms implement ward’s criterion?
\newblock \emph{Journal of classification}, 31\penalty0 (3):\penalty0 274--295,
  2014.

\bibitem[Matthey et~al.(2017)Matthey, Higgins, Hassabis, and
  Lerchner]{dsprites17}
Loic Matthey, Irina Higgins, Demis Hassabis, and Alexander Lerchner.
\newblock dsprites: Disentanglement testing sprites dataset.
\newblock https://github.com/deepmind/dsprites-dataset/, 2017.

\bibitem[Campbell et~al.(2015)Campbell, Veesler, Cheng, Potter, and
  Carragher]{campbell20152}
Melody~G Campbell, David Veesler, Anchi Cheng, Clinton~S Potter, and Bridget
  Carragher.
\newblock 2.8 {\aa} resolution reconstruction of the thermoplasma acidophilum
  20s proteasome using cryo-electron microscopy.
\newblock \emph{Elife}, 4:\penalty0 e06380, 2015.

\bibitem[Bepler et~al.(2019{\natexlab{b}})Bepler, Morin, Rapp, Brasch, Shapiro,
  Noble, and Berger]{bepler2019positive}
Tristan Bepler, Andrew Morin, Micah Rapp, Julia Brasch, Lawrence Shapiro,
  Alex~J Noble, and Bonnie Berger.
\newblock Positive-unlabeled convolutional neural networks for particle picking
  in cryo-electron micrographs.
\newblock \emph{Nature methods}, 16\penalty0 (11):\penalty0 1153--1160,
  2019{\natexlab{b}}.

\bibitem[Lin et~al.(2016)Lin, Zhu, Eng, Hudson, and Springer]{lin2016beta}
Fu-Yang Lin, Jianghai Zhu, Edward~T Eng, Nathan~E Hudson, and Timothy~A
  Springer.
\newblock $\beta$-subunit binding is sufficient for ligands to open the
  integrin $\alpha$iib$\beta$3 headpiece.
\newblock \emph{Journal of Biological Chemistry}, 291\penalty0 (9):\penalty0
  4537--4546, 2016.

\end{thebibliography}

\appendix

\section{Appendix}

\subsection{Calculating Kullback-Leibler divergence}
Based on the standard definition for the KL-divergence, we have:

\begin{align}
    \label{eq:app_kl_first}
    &KL(q(z,\theta,t,r|y)||p(z,\theta,t,r)) = \sum_{z,\theta,t,r}q(z,\theta,t,r|y)log\frac{q(z,\theta,t,r|y)}{p(z,\theta,t,r)}  \nonumber \\
    &= \sum_{z,\theta,t,r}q(t,r|y)q(\theta|t,r,y)q(z|t,r,y)log\frac{q(t,r|y)q(\theta|t,r,y)q(z|t,r,y)}{p(t,r)p(\theta)p(z)} \nonumber \\
    &= \sum_{t,r}q(t,r|y)log\frac{q(t,r|y)}{p(t,r)} \hspace{0.1cm}+  \nonumber \\ &\hspace{0.2cm}\sum_{z,\theta,t,r}q(t,r|y)q(\theta|t,r,y)q(z|t,r,y)log\frac{q(\theta|t,r,y)q(z|t,r,y)}{p(\theta)p(z)} 
\end{align}
To simplify this equation, we define the first part of the result from Equation \ref{eq:app_kl_first} as:
\begin{equation}
\label{eq:app_kl_t_r}
    KL_{t,r} = \sum_{t,r}q(t,r|y)log\frac{q(t,r|y)}{p(t,r)}
\end{equation}
The second part of the result from Equation \ref{eq:app_kl_first} can be further expanded as:
\begin{align}
    \label{eq:app_kl_theta_z_expansion}
    &\sum_{z,\theta,t,r}q(t,r|y)q(\theta|t,r,y)q(z|t,r,y)log\frac{q(\theta|t,r,y)q(z|t,r,y)}{p(\theta)p(z)} \nonumber \\
    &= \sum_{t,r} q\left(t,r|y\right) \left( \sum_{\theta} q\left(\theta|t,r,y\right)log\frac{q\left(\theta|t,r,y\right)}{p\left(\theta\right)} + \sum_{z}q(z|t,r,y)log\frac{q\left(z|t,r,y\right)}{p\left(z\right)} \right)  \nonumber \\
    &= \sum_{t,r} q\left(t,r|y\right) \left(KL\left(q\left(\theta|t,r,y\right)||p\left(\theta\right)  \right) + KL\left(q\left(z|t,r,y\right)||p\left(z\right) \right)   \right) 
\end{align}
Assuming these definitions for $KL_{\theta}$ and $KL_{z}$:
\begin{align}
    \label{eq:app_kl_theta}
    KL_{\theta} = KL\left(q\left(\theta|t,r,y\right)||p\left(\theta\right) \right) 
\end{align}
\begin{align}
    \label{eq:app_kl_z}
    KL_{z} = KL\left(q\left(z|t,r,y\right)||p\left(z\right) \right) 
\end{align}

, we can rewrite the the Equation \ref{eq:app_kl_first} using Equations \ref{eq:app_kl_t_r}, \ref{eq:app_kl_theta_z_expansion}, \ref{eq:app_kl_theta}, and \ref{eq:app_kl_z} as:
\begin{equation}
    \label{eq:app_kl_final}
    KL(q(z,\theta,t,r|y)||p(z,\theta,t,r)) = KL_{t,r} + \sum_{t,r} q\left(t,r|y\right) \left(KL_{\theta} + KL_{z} \right)
\end{equation}

\subsection{MNIST(N) and MNIST(U) datasets}

\begin{figure}[ht]
    \centering
    \includegraphics[width=\textwidth]{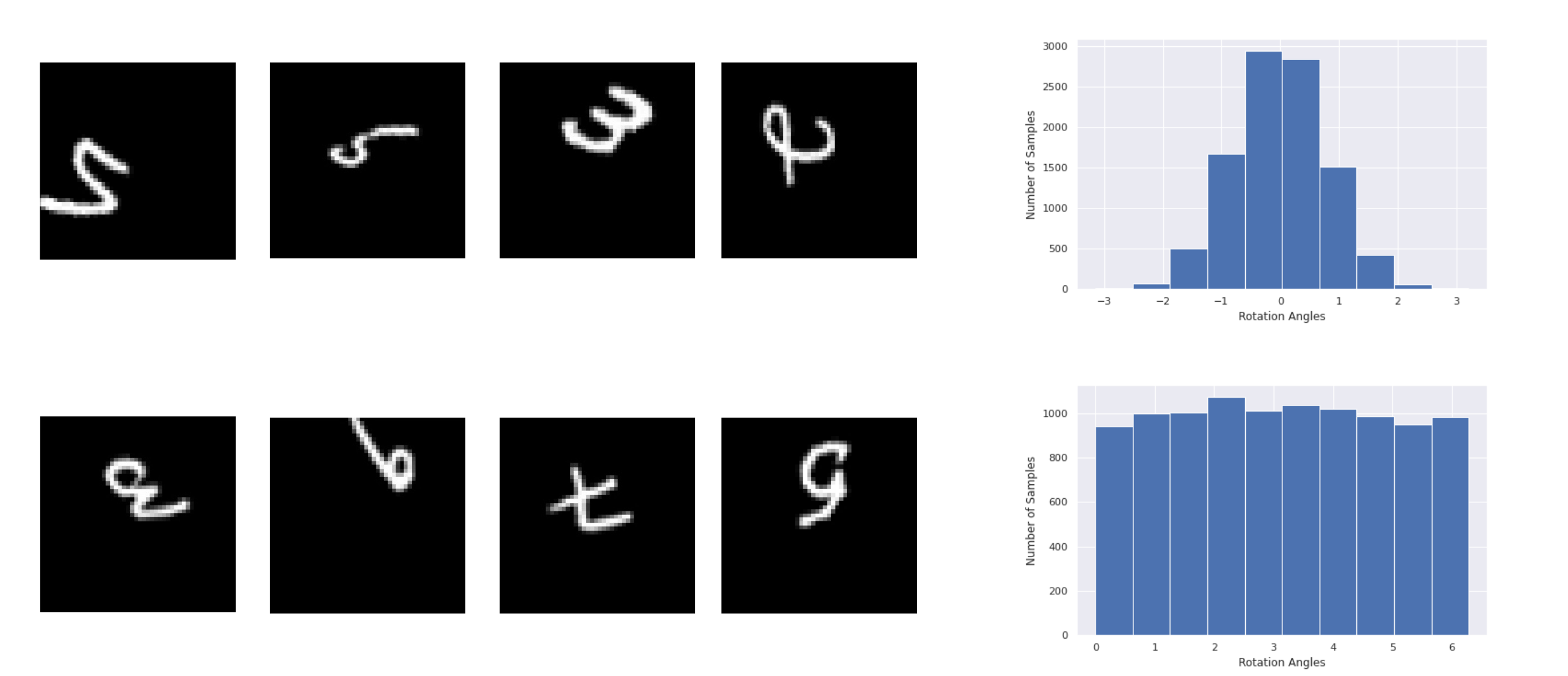}
    \caption{Left: Instances of MNIST(N) (top), and MNIST(U) (bottom) datasets. Right: Distribution of rotation angles in the test set of MNIST(N) (top), and MNIST(U) (bottom) datasets.}
    \label{fig:app_mnist_n_u}
\end{figure}

We generated two datasets of MNIST(N) and MNIST(U), by rotating and translating digits in MNIST. The rotation angles of digits for MNIST(N) are randomly sampled from $\mathcal{N}(0, \frac{\pi^{2}}{16})$, and for MNIST(U) are randomly sampled from \textit{U}(0, 2$\pi$) (Figure \ref{fig:app_mnist_n_u}). Images in both of the datasets are 50x50 pixels.

\subsection{Digit-wise rotation correlation, and RMSE of the predicted rotations}
\begin{figure}[ht]
  \centering
    \includegraphics[width=\textwidth]{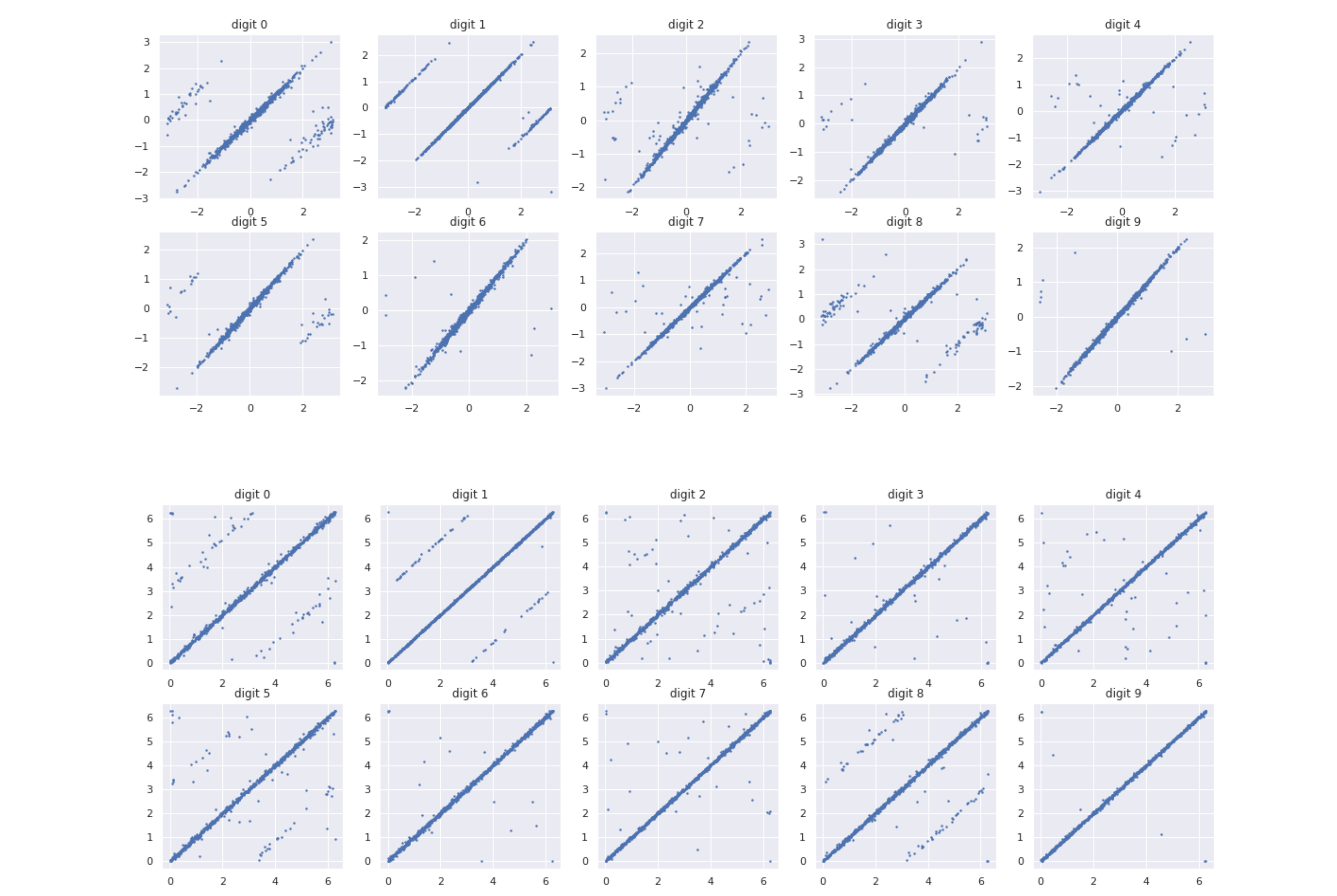}
    \caption{The predicted rotation values (x axis) and the ground-truth rotation angles (y axis) for digits in MNIST(N) (top), and MNIST(U) (bottom) using TARGET-VAE with model P$_8$ and z\_dim=2. Some predicted rotations for digits 0, 1, and 8 are off by $\pi$ from their ground-truth values.}
    \label{fig:app_mnist_digitwise}
\end{figure}

Following our discussion about the difference between the predicted angles and the ground-truth values, we measure the digit-wise rotation correlation for each dataset (Figure \ref{fig:app_mnist_digitwise}). We recognized that some predicted rotations are off by about $\pi$ from their ground-truth angles for digits 0, 1, and 8. We suspect that this is caused by the symmetry of these digits where for example a hand-written digit 8 with rotation $\pi$, looks roughly the same as that digit with rotation zero.

To study the accuracy of the predicted rotation angles by TARGET-VAE, we calculate the mean standard deviation of the predicted rotations, introduced in \cite{sun2021canonical}. This metric basically measures the mean square error between the rotation of the object in the input image and the predicted rotation for that object. We rotate each image in the test set of MNIST(U), 160 times using angles uniformly sampled from [0, 2$\pi$], and then pass them through the trained inference model to get the predicted rotation angle for them. Since assigning predicted rotation of zero to an object is arbitrary in our framework (model might assign predicted rotation angle of zero to the object that is actually rotated 90 degrees), we subtract the predicted rotation for the objects from the predicted rotation value when the input object is not rotated. Table \ref{tab:rotError}, shows the root mean square error (RMSE) calculated in average for each digit's predicted rotation. As expected, the RMSE of rotations is highest when there is no inference done on rotations as in spatial-VAE \cite{bepler2019explicitly}, and using finer discretization in rotation inference results in more accurate prediction of the rotation values. Some digits such as 0, 1, and 8, due to their symmetry, have less accurate rotation predictions compared to the other ones.

\begin{table}
  \caption{The RMSE of the predicted rotations over MNIST(U) with z-dim=2}
  \label{tab:rotError}
  \centering
  \resizebox{\textwidth}{!}{%
  \begin{tabular}{llllllllllll}
    \toprule
    Digits  & 0 & 1   & 2  & 3 & 4 & 5 & 6  & 7  & 8. & 9  & AVG\\
    \midrule
    Spatial-VAE \cite{bepler2019explicitly}  & 96.50 & 98.73   & 97.83  & 97.88  & 98.59  & 98.06   & 97.82  & 98.10  & 97.62  & 98.55  & 97.97\\ 
    TARGET-VAE P$_4$  & 31.57 & 33.86   & 32.36  & 11.66  & 27.08  & 20.94  & 6.26   & 8.15  & 28.95  & 6.68  & 20.75\\ 
    TARGET-VAE P$_8$  & 32.81 & 22.04   & 16.69  & 6.36  & 20.71  & 16.24  & 13.00   & 15.84  & 16.97  & 7.80  & 16.85\\ 
    TARGET-VAE P$_{16}$  & 17.39 & 18.47   & 14.74  & 5.79  & 7.65  & 12.09  & 3.40   & 4.63  & 17.75  & 2.74  & 10.47\\ 
    \bottomrule
  \end{tabular}
  }
\end{table}

\subsection{TARGET-VAE identifies multiple objects without supervision}
\begin{figure}[ht]
  \centering
    \includegraphics[width=\textwidth]{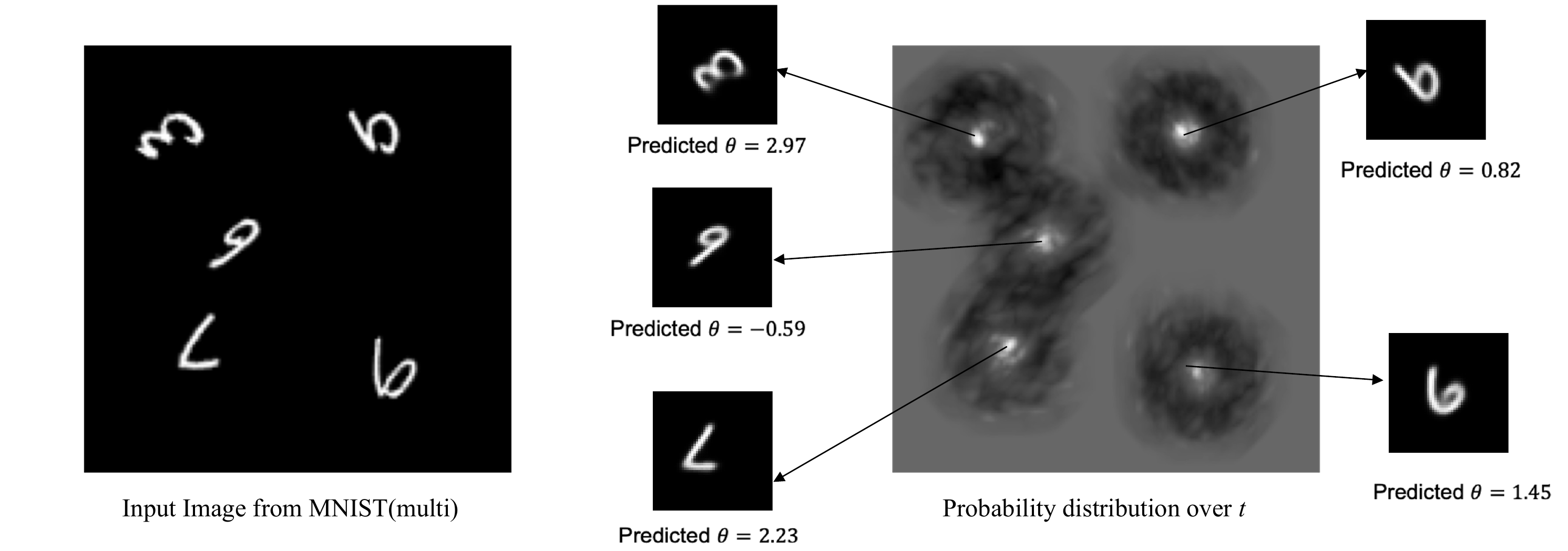}
    \caption{Left: Input image with multiple rotated and translated digits from MNIST(multi); Right: Probability distribution over $t$ ($q(t|y)$), the reconstructed objects and their predicted rotation angles. We marginalized $q(t,r|y)$ over $r$ to obtain $q(t|y)$ for visualization purposes. The high probability values in this attention map show the predicted locations of the objects. We use the peaks in $q(t,r|y)$ to sample from $q(\theta|t,r,y)$ and $q(z|t,r,y)$, the predicted rotation and content values for each object. The sampled $\theta$, $t$, and $z$ values are used to reconstruct each individual object.}
    \label{fig:mnist_multi}
\end{figure}

We created a new dataset using multiple rotated and translated digits from MNIST(U). We call this new dataset MNIST(multi) which we created by randomly sampling rotated and translated digits from MNIST(U) and inserting them in random offsets of a 150x150 pixel image. We use the model trained on MNIST(U) to identify the translation, rotation, and the content latent for the digits in MNIST(multi). We find that the model correctly identifies and reconstructs the objects (Figure \ref{fig:mnist_multi}). Even though the model, can identify non-overlapping objects in this approach, it struggles to identify the overlapping objects. We believe these preliminary results show the potential of our model to be used in multiple objects detection.

\subsection{TARGET-VAE predicts the rotation of the shapes in dSprites and clusters them with high accuracy}

\begin{figure}[ht]
  \centering
    \includegraphics[width=0.9\textwidth]{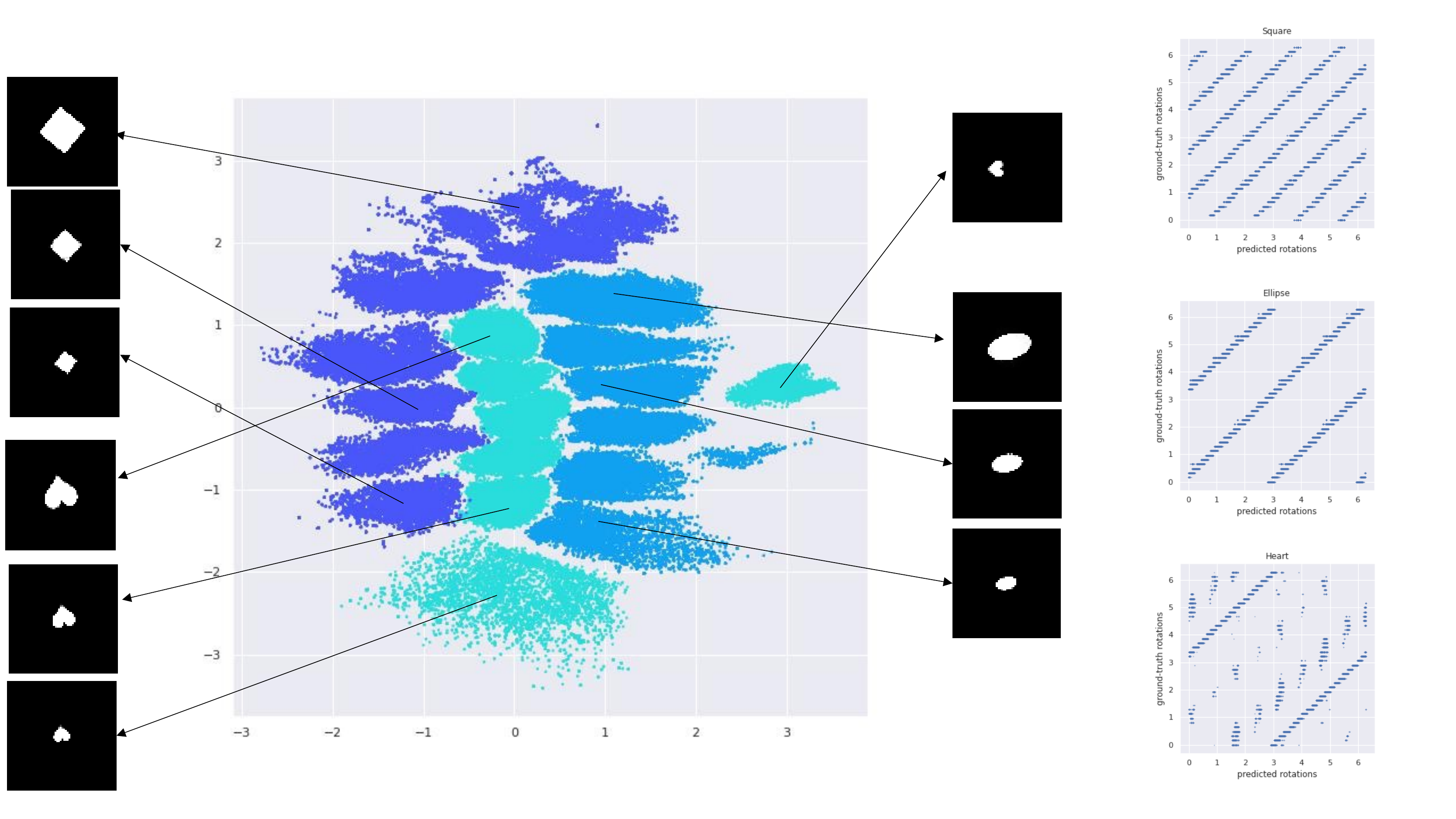}
    \caption{Left: Two-dimension latent space of dSprites using TARGET-VAE with P$_8$. Right: Correlation between rotations given by TARGET-VAE and ground truth rotations for squares, ellipses, and hearts.}
    \label{fig:dsprites_targetvae}
\end{figure}

The dSprites dataset was introduced to benchmark unsupervised disentangled representation learning methods \cite{dsprites17}. It contains 64x64 images of three shapes: squares, ellipses, and hearts. Each shape is rotated by one of 40 values linearly spaced in [0, 2$\pi$], translated across both $x$ and $y$ dimensions, and scaled using one of six linearly spaced values in [0.5, 1]. We train TARGET-VAE with P$_{8}$ group convolution and z\_dim=2 on dSprites, using a uniform prior over $r$ and $\mathcal{N}(0,\pi)$ prior over $\theta$. TARGET-VAE learns content representations that capture the type and scale of the objects, but is invariant from their location and rotation (Figure \ref{fig:dsprites_targetvae}). Furthermore, TARGET-VAE accurately predicts the rotation of the shapes, and, interestingly, the symmetry groups of the shapes become immediately apparent when examining the correlation between the predicted and ground truth rotations.

\subsection{TARGET-VAE learns transformation-invariant representations in the galaxy zoo dataset}
\begin{figure}[ht]
  \centering
    \includegraphics[width=0.9\textwidth]{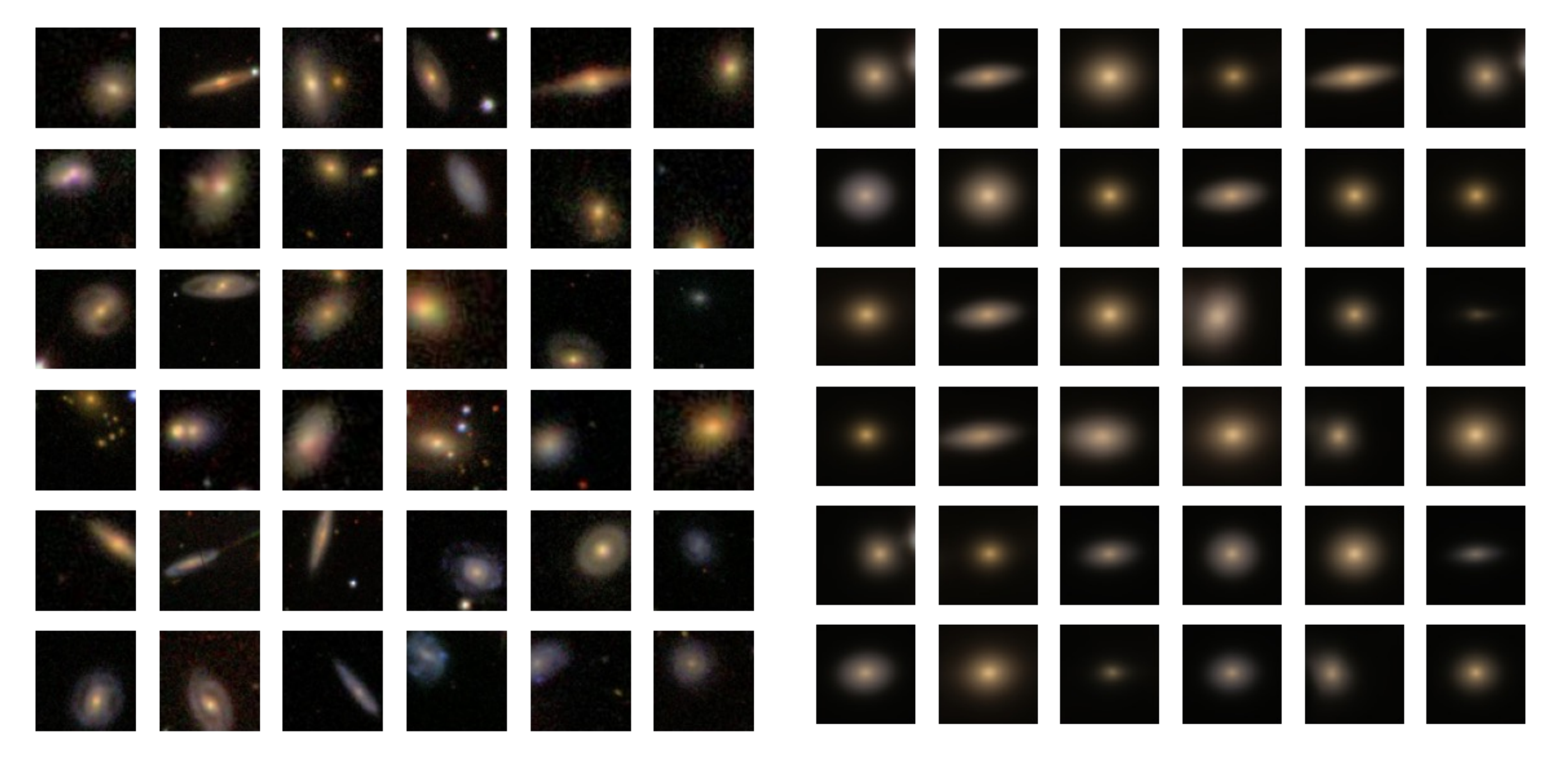}
    \caption{Left: Images of galaxy dataset used for testing. Right: Reconstructed images by TARGET-VAE with P$_8$ and z\_dim=2 and using the transformation-invariant representations, where rotation and translation are set to 0.}
    \label{fig:app_galaxy}
\end{figure}

Galaxy zoo contains images of galaxies gathered by Sloan Digital Sky Survey \cite{lintott2008galaxy}. This dataset contains more than 61,000 RGB images, which we cropped and downsampled to 64x64 pixels. The galaxies appear with different rotations and translations in the images. We train TARGET-VAE with P$_8$ group convolution on the train set. TARGET-VAE learns to accurately predict the rotation and translation of the galaxies and reconstructs centered and aligned galaxy images when generating with $t$ and $\theta$ set to zero (Figure \ref{fig:app_galaxy}).

\subsection{Ablation studies}
We conducted extensive ablation study to validate the effectiveness of the main components of our proposed method.

\paragraph{Variant 1 - Inference only on translation}

We evaluated the importance of the rotation equivariant features, by modifying the inference model to use regular convolutional layers instead of the group convolutional ones. As a result, the posterior distributions will be $q(t|y)$, $q(z|t,y)$, and $q(\theta|t,y)$, where they only depend on the input $y$ and the translation value $t$. We observed that, as expected, eliminating inference on the discretized rotation dimension has a significant negative effect on identifying transformation-invariant representations and the clustering accuracy on MNIST(U) is only 33.8\% (Table \ref{tab:app_ablation}).  

\paragraph{Variant 2 - Inference only on translation + using group convolutional layers}

In this variant, we use the group convolutional layers in the inference model, but we do not perform any inference on the rotation dimension. We apply a fully-connected layer on the output of the group convolutional layers to map the rotation values at every location to a single value. As a result, the posterior distribution are the same as variant 1 and they do not depend on $r$. We experiment with this variant to identify how effective is the use of the group convolution layers. It turns out that if we do not perform any inference on the rotation dimension, just using the group convolutional layers for feature extraction, only slightly improves the clustering accuracy. 

\paragraph{Variant 3 - Inference on both translation and rotation without adding $\theta_{offset}$:}

In this variant, we use group convolutional layers and we perform inference on both translation and rotation. The difference between this variant and our proposed framework is that we are not adding $\theta_{offset}$ of kernels in each rotation dimension to the $q(\theta|t,r,y)$. Adding $\theta_{offset}$ to the posterior on $\theta$, allows us to break down the rotation space among the $r$ discretized rotations, and without it, the model proves not capable of identifying the rotation of the digits in the MNIST(U) dataset.

\paragraph{Variants 4 to 6 - Increasing the level of discretization of the rotation space}

In these variants, we perform inference on both rotation and translation, and we add $\theta_{offset}$ of each rotation dimension to the mean of its corresponding q($\theta$|t,r,y) distribution. In these variants, we show that by increasing the level of discretization of the rotation space, model can have a better estimate of the actual rotation values, and this in turn helps the model with improving the clustering accuracy.

Table \ref{tab:app_ablation}, shows the translation and rotation correlation, along with the clustering accuracy of the mentioned variants on the MNIST(U) dataset. Performing inference on rotation plus adding $\theta_{offset}$ to the posterior on $\theta$, significantly increases the rotation correlation and the clustering accuracy. 

\begin{table}
  \caption{Performace of variants in the ablation study on MNIST(U)}
  \label{tab:app_ablation}
  \centering
  \begin{tabular}{lllll}
    \toprule
    Model  & Group Conv & Translation Corr   & Rotation Corr  & Clustering Accuracy   \\
    \midrule
    Variant 1   & - & 0.966, 0.967  & 0.005  & 33.8\% \\
    Variant 2   & p4 & 0.967, 0.967  & 0.005  & 37.1\% \\
    Variant 3   & p4 & 0.968, 0.972  & 0.008  & 36.6\% \\
    Variant 4   & p4 & 0.975, 0.976  & 0.80  & 56.6\% \\
    Variant 5   & p8 & 0.972, 0.971  & 0.859  & 57.1\% \\
    Variant 6   & p16 & 0.974, 0.971  & 0.93  & 63.4\% \\
    \bottomrule
  \end{tabular}
\end{table}

\subsection{Learning translation-rotation-invariant representations of proteins with TARGET-VAE}
In cryo-EM images, rotation and translation are the major transformations that cause variations in particles in the micrograph. Here, we show the result of our experiments with TARGET-VAE to learn the translation-rotation-invariant representations, for two cryo-EM datasets.

\paragraph{Identifying hinge motion of 5HDB}

\begin{figure}[ht]
  \centering
    \includegraphics[width=\textwidth]{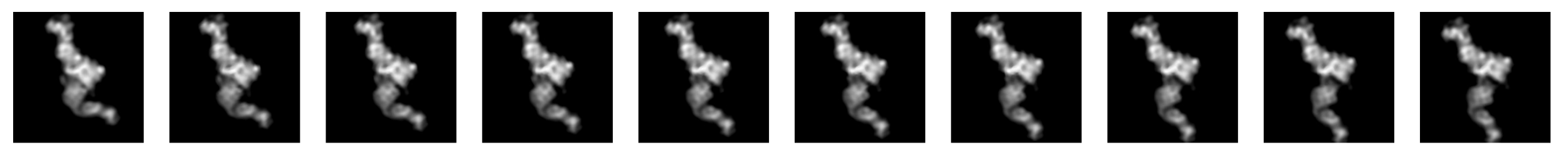}
    \caption{Reconstructed proteins from 5HDB dataset by sampling from 1D representation space, shows the movement in the lower part of the particle.}
    \label{fig:app_5hdb}
\end{figure}

We train our model on a dataset of 20,000 simulated projections of integrin $\alpha$-IIb in complex with integrin $\beta$-3 (5HDB) \cite{lin2016beta}. We aim to identify the translation and rotation invariant representations of the protein to be able to identify the variations in its structure. We train TARGET-VAE with P$_8$ and uniform prior over $\theta$. Since there is less variation in the data, we set $z\_dim$ to $1$. After training, we sample from the representation latent space and reconstruct the images with no rotation and translation. Figure \ref{fig:app_5hdb} shows some examples of the reconstructed particles. We observe that the reconstructed images identify the hinge motion of the particle.

\paragraph{Learning arm motion of particle in CODH/ACS}

\begin{figure}[ht]
  \centering
    \includegraphics[width=\textwidth]{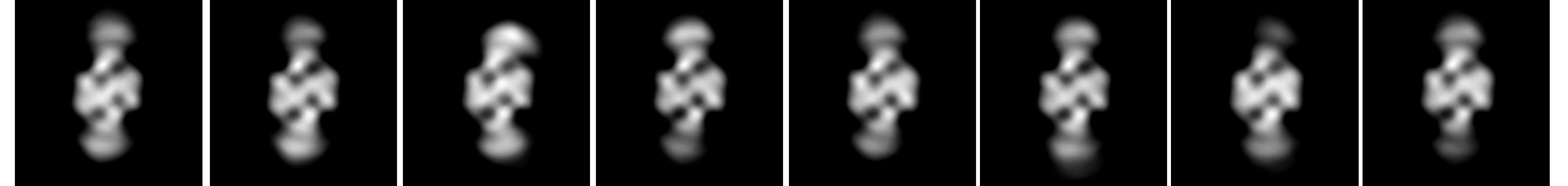}
    \caption{Reconstructed proteins from CODH/ACS dataset, where the motion in the upper and lower arms of the particle can be captured.}
    \label{fig:app_codhacs}
\end{figure}

We have about 14,000 40x40 pixels images of the CODH/ACS protein complex. We train TARGET-VAE with P$_8$ and $z\_dim=2$, for 100 epochs. The prior over $\theta$ is uniform and we set the generator to have 6 fully-connected layers with 512 hidden units in each. After training, we sample from the transformation-invariant representation space and reconstruct the particle to identify the different conformations of the protein. Figure \ref{fig:app_codhacs} shows the movement on the upper and lower parts of the particle.

\end{document}